\documentclass[10pt,twocolumn,letterpaper]{article}
\usepackage{times}
\usepackage{epsfig}
\usepackage{graphicx}
\usepackage{amsmath}
\usepackage{amssymb}
\usepackage{color}

\usepackage{comment}

\usepackage{flushend}

\usepackage{amssymb}
\usepackage{bm}

\usepackage{subcaption}
\usepackage{enumitem}

\usepackage{algorithm}
\usepackage[noend]{algpseudocode}

\usepackage[font=small,skip=0pt]{caption}

\usepackage[T1]{fontenc}
\usepackage[utf8]{inputenc}
\usepackage{authblk}

\begin{document}

\title{ KNN-enhanced Deep Learning Against Noisy Labels} % Replace with your title
\author[1]{Shuyu Kong\thanks{shuyukong2020@u.northwestern.edu}}
\author[1]{You Li \thanks{you.li@u.northwestern.edu }}
\author[2]{Jia Wang\thanks{jwang34@iit.edu }}
\author[3]{Amin Rezaei\thanks{me@aminrezaei.com }}
\author[1]{Hai Zhou\thanks{haizhou@northwestern.edu }}
\affil[1]{Department of Electrical and Computer Engineering, Northwestern University}
\affil[2]{Department of Electrical and Computer Engineering, Illinois Institute of Technology}
\affil[3]{Department of Computer Engineering and Computer Science, California State University Long Beach}
\date{\vspace{-5ex}}

\maketitle
%%%%%%%%% ABSTRACT
\begin{abstract}
Supervised learning on Deep Neural Networks (DNNs) is data hungry. Optimizing performance of DNN in the presence of noisy labels has become of paramount importance since collecting a large dataset will usually bring in noisy labels. Inspired by the robustness of K-Nearest Neighbors (KNN) against data noise, in this work, we propose to apply deep KNN for label cleanup. Our approach leverages DNNs for feature extraction and KNN for ground-truth label inference. We iteratively train the neural network and update labels to simultaneously proceed towards higher label recovery rate and better classification performance. Experiment results show that under the same setting, our approach outperforms existing label correction methods and achieves better accuracy on multiple datasets, e.g., $76.78\%$ on Clothing1M dataset.

%Deep neural networks nowadays are data-hungry. Collecting a large dataset will usually bring in noisy labels. In this work, we assume the most general noisy label configuration: there is no clean reference dataset, no transition relations of noisy labels between classes, and no available distribution of true labels.

%Our approach leverages deep neural networks for feature extraction and \textit{k}-nearest neighbor for label correction. Feature vectors are extracted for samples by simulating the current deep neural network and taking the values from an intermediate layer. Then those vectors are compressed and put into \textit{k}-nearest neighbor to infer their true labels. As training proceeds, the loss function will generally shift its weight from the original labels to the labels inferred from \textit{k}-nearest neighbor in order to avoid overfitting. 

%Deep learning has achieved great success in many vision applications. Supervised classification is most common problem where deep learning is demonstrated to outperform traditional approaches. However, on limitation of deep learning is 

%Noisy labels, using KNN, experiments: Mnist, Fmnist, Cifar10, Cifar100, SVHN.

%Use multiple layers as embedding, extra unsupervised reconstruction loss to help handle high noisy label.

%The ubiquitous use of deep architectures in different areas of computer vision, 
\end{abstract}

%%%%%%%%% BODY TEXT
\section{Introduction}
Deep Neural Networks (DNNs) have achieved remarkable success in various applications including computer vision, speech recognition, and robotics. Supervised learning on DNNs is data hungry. Obtaining a large dataset with labels at an affordable cost is usually done by crowdsourcing~\cite{yan2014learning,zhang2016understanding} and web query~\cite{schroff2010harvesting,yang2018recognition}. Each of those would inevitably introduce a significant amount of noisy labels. On the other hand, DNNs are prone to overfit noisy training data~\cite{zhang2016understanding,arpit2017closer}, and their generalization performance is downgraded as a result.

To resist noisy labels in training DNNs, numerous methods have been proposed, including robust loss formulation~\cite{patrini2017making,goldberger2016training,Wang_2019_ICCV}, curriculum learning~\cite{Hacohen_2019_CL} and label correction~\cite{Yi_2019_CVPR,Tanaka_2018_CVPR}. In this paper, we focus on label correction approach which alternatively sanitizes noisy labels and improves the model performance. Previous work leverages prediction from DNN itself to infer the ground truth labels~\cite{Yi_2019_CVPR,Tanaka_2018_CVPR}. However, such prediction is likely to be poisoned by the noise in the training dataset. Therefore, we are motivated to seek for a more robust label correction approach.

In this paper, we leverage the deep K-Nearest Neighbors (KNN) algorithm to facilitate learning with noises. The KNN algorithm assumes that similar things exist in close proximity. KNN is a favorable classification approach when no prior knowledge on sample distribution are available, and has shown robustness against adversarial examples ~\cite{Sitawarin_2019,papernot2018deep,wang2019evaluating}. Our approach is based on a key observation that during the learning phase, useful features are learnt in the intermediate layers despite the presence of corrupted labels in the dataset. We propose to use those features to discover similarity among samples. Even though the final labels are different for two samples belonging to same category, their features share high similarity.

%\{Hai: Perhaps we should discuss that the difference will increase with the layer getting closer to the output. The tricky question is why not use the original input.\}

%https://arxiv.org/pdf/1908.06112.pdf

%\color{red} Is this the disadvantage of using KNN? 
%Intuition is as training going on, when the labels are corrected, the model already overfits training noise, which is hard to recover. There are two remedies, one is to reset the learning rate, the other more nature option is to retrain the neural network from scratch.
%\color{black}

Overall, we present a framework that iteratively applies deep KNN to infer ground truth labels and retrains the neural network with the predicted labels, thus simultaneously making progress toward higher label recovery and better classification performance. It is a generalized framework that does not require an estimation of noise transition distributions or a clean dataset for reference. We also propose two KNN label correction algorithms. The first one, IterKNN, uses all the labels to infer ground truth labels. The second one, SelKNN, selects a certain amount of clean examples as reference for KNN ground truth inference. The selection principle is to find samples with small cumulative normalized loss because those samples are more likely to have correct labels. We empirically show that our approach achieves state-of-art performance.
%With symmetric cross entropy loss, our approach can achieve overall better accuracy than several existing approaches.

%Many approaches update labels based on model's prediction(self knowledge distillation). We empirically show KNN's prediction is more robust than model itself's prediction.

The contributions of this paper are as follows:
\begin{itemize}[topsep=0pt,itemsep=-1ex,partopsep=1ex,parsep=1ex]
	\item To our best knowledge, we are the first to apply deep KNN for label correction in corrupted training dataset. The features are extracted from intermediate layers of neural network. To further mitigate the impact of noisy labels, we propose a loss ranking approach to select samples with high confidence to be labelled correctly as reference for KNN prediction.
	
    \item We explore the benefits of iterative retraining after label correction. We show that iterative retraining can help neural network escape from over-fitting and discover more corrupted labels, thus achieves better performance.
    
    \item We conduct extensive experiments to demonstrate the robustness of deep KNN against label noise. We found out that even though the final prediction is corrupted by the noise in training dataset, CNN can still learn robust and useful features in deep layers to facilitate KNN for ground-truth label inference. We also provide insights on how deep feature is better than both of the shallow features and final logits with regard to noisy label correction..
\end{itemize}

%https://papers.nips.cc/paper/8072-co-teaching-robust-training-of-deep-neural-networks-with-extremely-noisy-labels.pdf

%https://openreview.net/pdf?id=BklIxyHKDr

%https://openreview.net/pdf?id=S1xnKi5BOV

%https://arxiv.org/pdf/1905.05040.pdf

%https://github.com/subeeshvasu/Awesome-Learning-with-Label-Noise
\section{Related Work}

\subsection{Generalization of Deep Neural Networks}

Zhang \textit{et al.}~\cite{zhang2016understanding} showed that deep neural networks have the capacity to memorize completely random labels, but may result in poor generalization. This indicates that label corruption in training dataset has a negative impact on DNN performance. Other studies~\cite{DBLP:journals/corr/abs-1903-11680} demonstrated that DNN tends to learn clean labels first and that overfitting to corrupted labels requires to stray far from initialization. Thus early stop is a simple but effective method to resist noisy labels.

\subsection{Semi-Supervised Deep Learning}
The goal of semi-supervised learning is to learn from partially labelled dataset. With the development of DNNs, researchers have studied how DNN can learn in semi-supervised setting. One methodology is to add an unsupervised loss term as regularizer to force mutual exclusiveness of different classes~\cite{tarvainen2017mean,Sajjadi_2016}. Another popular approach in semi-supervised deep learning is to assign pseudo-labels to unlabeled examples. The pseudo-labeled data are trained in the supervised fashion. Ahmet \textit{et al.}~\cite{Iscen_2019_CVPR} have proposed transductive label propagation using nearest neighbor graph. Our approach can be viewed as label propagation from clean examples to noisy examples. However, our problem is more challenging because we do not assume that clean examples are provided. 

\subsection{Noisy Label Learning}
Loss correction approach is widely studied in training with noisy labelled data. Forward backward loss correction~\cite{patrini2017making} directly leverages the noisy transition matrix $T$ to modify the Cross Entropy (CE) loss. But in practice $T$ is usually not given. Other studies attempt to estimate the noisy transition matrix by modelling it with a fully connected layer~\cite{sukhbaatar2014training,Jindal_2016}, or use bootstrapping to avoid direct noise modelling ~\cite{reed2014training}. Recently, Wang \textit{et al.}~\cite{Wang_2019_ICCV} have proposed to combine CE with Reverse Cross Entropy (RCE) and demonstrated that Symmetric Learning (SL) can avoid overfitting noisy data. 

The field of Curriculum Learning (CL), which is motivated by the idea of a curriculum in human learning, attempts at imposing some structure on the training set. Such structure essentially relies on a notion of ``easy'' and ``hard'' examples, and utilizes this distinction in order to teach the learner how to generalize easier examples before harder ones \cite{Hacohen_2019_CL}. Empirically, the use of CL has been shown to accelerate and improve the learning process \cite{Bengio_2009_CL} and noise robustness \cite{Braun_2017_CL}. 

%SL is empirically shown to achieve state-of-art performance on multiple datasets with different noise levels.

Noisy label detection and filtering is another approach to mitigate data noise~\cite{Huang_2019_ICCV}. This is based on the fact that removing corrupted data can improve model performance. However, the hard samples may be confused with the noisy ones. As a consequence, certain amount of hard samples are filtered out together with noisy samples. The resultant clean dataset contains most of the easy samples, which will make the classifier less generalizable.

Other methods attempt to correct noisy labels. Tanaka \textit{et al.}~\cite{Tanaka_2018_CVPR} have formulated the noisy label learning problem as a joint optimization problem where model parameters and data sample labels are optimized alternatively. Yi \textit{et al.}~\cite{Yi_2019_CVPR} also have targeted the optimization problem by concurrently updating noisy labels and model parameter. Both methods require the prior knowledge over label distribution in order to be effective. However, in practice it is difficult to obtain such prior knowledge.

\subsection{Robust K-Nearest Neighbors}
KNN is a favorable classification approach when no prior knowledge on sample distribution is available, and is believed to be resistant to label noise. Gao \textit{et al.} \cite{gao2016resistance} have provided theoretical analysis on the robustness of KNN against noisy labels for binary classification. For symmetric noise where the labels of each class have equal probability to be flipped to the other, they showed that KNN can reach the same consistent rate as in the noise free setting. As for asymmetric noise, KNN is still robust since most samples can be correctly classified. They then derived a strategy to deal with noisy label classification with KNN. Specifically, their method corrects the labels of ``totally misled'' samples. How the labels should be corrected is depending on the estimations of noise proportions~\cite{liu2015classification,menon2015learning}.

Another strategy to resist label noise with KNN is to emphasize the more trustworthy information. Parvin \textit{et al.}~\cite{parvin2010modification} assign weights to every sample based on its validity. On the other hand, it is shown that to assign weights to features can also improve robustness and accuracy for KNN~\cite{itqon2001improving}.

An earlier work~\cite{bahri2020deep} proposes to use KNN for selecting clean examples. It shows that a simple KNN filtering approach on the logit layer of a preliminary model can remove mislabeled training data and help produce more accurate models. This work is most closely related to ours. But we want to highlight several difference between the two works and address our novel contribution on combining deep KNN inference with noisy learning. Firstly, instead of only filtering out suspicious examples, our work attempts to use KNN to correct corrupted labels so that the noisy examples are recycled to augment the training dataset, thus boosting the generalization performance of the final model. Besides, we also study the effect of selecting different layer as features on the KNN correction performance rather than restricting choice of only extracting the last logit layer. Last but not least, we discover the disadvantage of adopting the whole training dataset as KNN reference (as done in IterKNN) when the noisy level is high. Therefore, we propose a clean example selection strategy (as done in SelKNN) based on ranking the cumulative training loss with the hope that noisy labels will be mostly corrected by clean labels as reference for KNN classification.

\begin{figure*}
\setlength{\belowcaptionskip}{-0.3cm}
\begin{center}
%\fbox{\rule{0pt}{2in} \rule{.9\linewidth}{0pt}}
\includegraphics[width=0.9\textwidth]{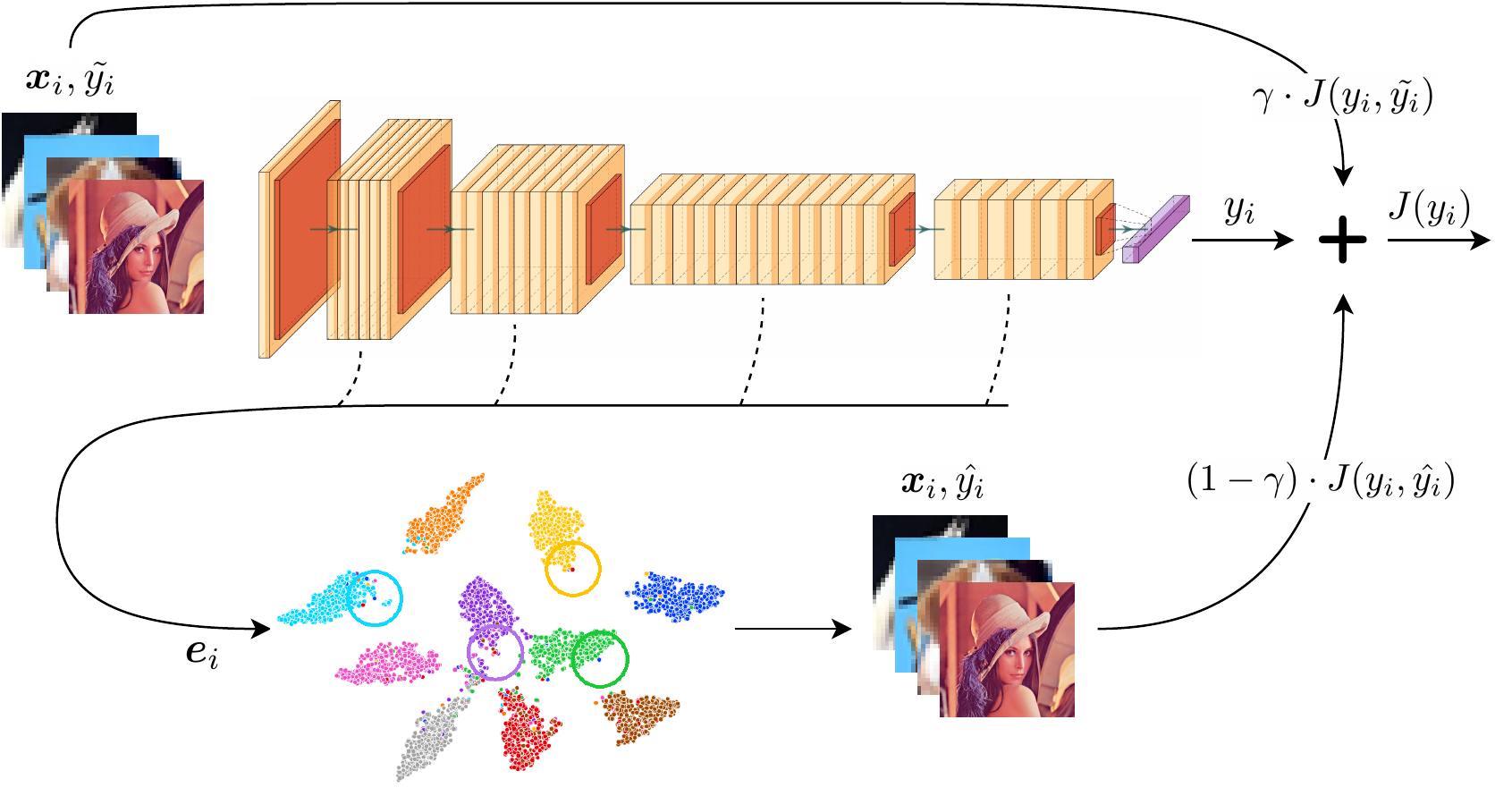}
\end{center}
    
    \caption{The overall architecture of IterKNN. The deep neural network extracts embeddings of samples with intermediate layers. After every training episode, KNN classifier corrects labels in the reference dataset based on the embeddings of the samples. The deep neural network is trained on a hybrid loss function, which is comprised of the loss on the original labels and the loss on KNN predicted labels.}
\label{fig:short}
\end{figure*} 
\section{Main Approach}

\subsection{Preliminaries and Problem Statement}

%We are targeting a multiclass classification problem with label noise. Let $\mathcal{X}$ and $\mathcal{Y}$ denote the instance space and label space, respectively. Let $\tilde{S_n} = (\bm{x}_1, \tilde{y_1}),  (\bm{x}_2, \tilde{y_2}), \cdots,  (\bm{x}_n, \tilde{y_n})$ be the training dataset $D$. We assume a random noise model, in which we can not observe the true label $y_i\ (i \in n)$ for every sample. Each true label $y_i$ has a probability to be flipped to any label $\tilde{y_i}\ (i \in n)$ based on a noise transition matrix $T$. The overall label error rate on the dataset due to noise is defined as $Pr_{\tilde{y_1}\sim\mathcal{Y}}(\tilde{y_i} \neq y_i)$.

%We make an assumption that there exist a latent space, $\mathcal{L}$, in which instances close to each other have similar labels. $f: \mathcal{X} \rightarrow \mathcal{L}$ be the transformation with desired properties: it maps instances to a space that has continuity and with succinct representation. In this work, we use a deep neural network with parameter $\theta$, and choose $f$ to be the first several layers of the network. On space $\mathcal{L}$, label $\tilde{y_1}$ is corrected to $\tilde{y_1'}$ according to a \textit{k-nearest neighbor} classifier by taking the majority of label classes: $\eta(x) = arg\ max_y\ \eta_k(y:x)$, where $\eta_k(y:x)$ is the discriminant function for k nearest neighbor.

We are targeting a multiclass classification problem with label noise. Let $\mathcal{X}$ and $\mathcal{Y}$ denote the image space and label space, respectively. Let $\tilde{D_n} = \{(\bm{x}_1, \tilde{y_1}), \cdots,  (\bm{x}_n, \tilde{y_n}) \}$ be the noisy training dataset where $x_i\in \mathcal{R}^{d}$ is a d-dimensional vector and $\tilde{y_i}\in \mathcal{Y}=\{1,...,C\}$. True labels $y_i,\cdots,y_n$ are not observable. There exists a hidden noise model $T$, while $t_{i} = P(\tilde{y_i} | x_i, y_i)$ representing the probability the true label $y_i$ of instance $x_i$ is flipped to $\tilde{y_i}$. Note if $T$ is only class-dependent, we will have $t_i = P(\tilde{y_i} | y_i)$, which is assumed in some existing work~\cite{patrini2017making}. Consequently, the overall label error rate on the dataset due to noise is defined as $Pr(\tilde{y_i} \neq y_i)$. 
%$Pr_{( y_i, \tilde{y_i}) \sim ( \mathcal{Y}, \tilde{\mathcal{Y}})}(\tilde{y_i} \neq y_i)$. 

When the dataset is clean, the learning problem can be formulated into an optimization problem as follows:

\begin{equation} \label{clean formulation}
    \mathcal{\theta^*} = \underset{\theta}{\mathrm{argmin}}\dfrac{1}{n} \sum^{n}_{i=1} J(y_i, \mathcal{F}(x_i, \theta)),
\end{equation}

where $\mathcal{F}$ is a classifier with parameters $\theta$ and $J$ is the target loss function. The objective is to find optimal parameters $\theta^*$ such that average loss is minimized. However, when labels are corrupted, we can not obtain the above formulation since ground truth labels $y_i,\cdots,y_n$ are unknown. Therefore, only corrupted labels can be used in the following optimization problem:
\begin{equation}
    \mathcal{\tilde{\theta^*}} = \underset{\theta}{\mathrm{argmin}}\dfrac{1}{n} \sum^{n}_{i=1} J(\tilde{y_i}, \mathcal{F}(x_i, \theta)).
\end{equation}

$\mathcal{\tilde{\theta^*}}$ is only a sub-optimal solution to Equation \ref{clean formulation}. To approach the real optimal solution $\theta$, we attempt to detect and correct the corrupted labels based on the following fact:
\begin{equation}
     \mathcal{\theta^*} = \underset{y\rightarrow \tilde{y}}{\mathrm{lim}} \mathcal{\tilde{\theta^*}}.
\end{equation}

It is also important to develop a method agnostic to $T$, that is, we neither assume any structure nor require any prior knowledge of noise model $T$.

\subsection{Deep \textit{k}-Nearest Neighbor Label Correction}

\textit{k}-Nearest Neighbor is a widely used non-parametric classification approach. It predicts the label of input by finding a total of \textit{k} nearest neighbors and then taking a majority voting of the labels of those neighbors:
\begin{equation}
    \eta_k(\bm{x}) = \underset{y}{\mathrm{argmax}} \sum_i^n 1\cdot(y=y_i, \bm{x}\in N_k(\bm{x}_i)),
\end{equation}

in which $\eta_k(\bm{x})$ is the label prediction for sample $\bm{x}$, and $N_k(\bm{x})$ is some distance matrix.

Inspired by the robustness of KNN against noisy dataset~\cite{gao2016resistance}, we propose to apply KNN on label correction. It requires to define a distance metric to measure the similarity between image samples. One straightforward solution is to compute the total pixel-wise difference between pairs of images. Yet such a metric is weak even toward a very small transformation. The most popular approach is to measure the distance in feature space. Consequently, one critical question is what kind of features we should use to represent each image.

Traditional offline feature extraction approaches like SIFT~\cite{lowe2004distinctive} and HOG~\cite{dalal2005histograms}
can resist transformations, but may not preserve enough class-related information for classification. As deep learning becomes dominant in vision-related problems, deep feature extraction attracts more and more attention and shows state-of-art performance in downstream tasks. In this paper, we use the deep representation learned by the neural network as features. More specifically, we take the intermediate layers from the neural network being trained as feature extractors. Even though the training dataset is corrupted, the deep layers of DNN can still extract useful features (embeddings). We will empirically show and validate such a result in Sec.~\ref{Sec: KNN robustness}. Moreover, as noisy labels get corrected by KNN in every episode, the DNN can be trained on more accurate data and thus produces features with higher quality. The better features can further facilitate KNN for more accurate label correction. Such a process forms a benign cycle, and benefits both label correction and neural network training as the result.

%The intuition is that DNN can extract useful features or embeddings even if trained from a corrupted dataset, which we will empirically show and validate in Section~\ref{}.

%self-learning or self knowledge distillation

\makeatletter
\def\BState{\State\hskip-\ALG@thistlm}
\makeatother
\begin{algorithm}
\setlength{\belowcaptionskip}{-0.3cm}
\caption{Iterative Training and Correction Algorithm}\label{Alg:Iter}
\hspace*{\algorithmicindent} \textbf{Input:} the noisy training set $\tilde{S_n}$, initial network model $\theta_0$, $k$, number of training episodes \textit{M}, number of training epochs \textit{T}, initial learning rate $\epsilon_0$, loss coefficient $\gamma$.
\begin{algorithmic}[1]\\

%\Procedure{cyclic training with KNN label correction}{}\\
%\Input{Noisy Training Set $\tilde{S_n}$, $k$, number of epochs \textit{E}}\\
%$\hat{S_n}\leftarrow \tilde{S_n}$\\
Initialize KNN corrected set $\hat{S_n}$ to noisy set $\tilde{S_n}$\\
\textbf{for} \textit{m} $\leftarrow$ 1, \textit{M} \textbf{do}\\
\hskip1em (re)Initialize $\theta$ to $\theta_0$\\
\hskip1em (re)Initialize $\epsilon$ to $\epsilon_0$\\
%\hskip1em $\theta \leftarrow \theta_0$\\
\hskip1em \textbf{for} \textit{t} $\leftarrow$ 1, \textit{T} \textbf{do}\\
\hskip2em Update $\theta$ at training rate $\epsilon$ with loss function: 

\hskip1em $(1-\gamma)\cdot J(\theta; \bm{x}, \hat{y}) + \gamma \cdot J(\theta; \bm{x}, \tilde{y})$\\
\hskip2em Update learning rate $\epsilon$ accordingly\\
\hskip1em \textbf{end}\\
\hskip1em $\hat{S_n}\leftarrow$ label\_correction($\hat{S_n}$)  // either IterKNN or SelKNN\\
\hskip1em Decay loss coefficient $\gamma$\\
\textbf{end}\\
$S_n^* \leftarrow \hat{S_n}$\\
Train $\theta$ by training neural network on $\hat{S_n}$
%\EndProcedure
\end{algorithmic}
\hspace*{\algorithmicindent} \textbf{Output:} corrected set $S_n^*$, trained network model $\theta$.
\end{algorithm}

\begin{comment}
\begin{algorithm}
\setlength{\belowcaptionskip}{-0.3cm}
\caption{Iterative Training and Correction Algorithm}\label{Alg:Iter}
\hspace*{\algorithmicindent} \textbf{Input:} Noisy Training Set $\tilde{S_n}$, $k$, number of epochs \textit{T}
\begin{algorithmic}[1]\\

%\Procedure{cyclic training with KNN label correction}{}\\
%\Input{Noisy Training Set $\tilde{S_n}$, $k$, number of epochs \textit{E}}\\
\textbf{for} i $\leftarrow$ 1 \textbf{to} T \textbf{do}\\
\hskip1em (Re)initialize $\theta$\\
\hskip1em Train $\theta$ by training neural network on $\tilde{S_n}$\\
\hskip1em $E$ = $\{\}$\\
\hskip1em \textbf{for} $\bm{x}_i, \tilde{y_i}$ in $\tilde{S_n}$ \textbf{do}\\
\hskip2em Simulate embedding $\bm{e}_i$ with $f(\theta, \bm{x}_i)$\\
\hskip2em $E$ = $E\ \cup\ (\bm{e}_i, \tilde{y_i})$\\
\hskip1em \textbf{end}\\
\hskip1em \textbf{for} $\bm{e}_i, \tilde{y_i}$ in $E$ \textbf{do}\\
\hskip2em Compute $\tilde{y_i'}$ by $k$-nearest neighbor\\
\hskip2em Update $\tilde{y_i}$ with $\tilde{y_i'}$ in $\tilde{S_n}$\\
\hskip1em \textbf{end}\\
\textbf{end}\\
$S_n^*$\ = \ $\tilde{S_n}$\\
Train $\theta$ by training neural network on $\tilde{S_n}$
%\EndProcedure
\end{algorithmic}
\hspace*{\algorithmicindent} \textbf{Output:} Corrected Set $S_n^*$, trained network model $\theta$.
\end{algorithm}
\end{comment}

\subsection{Iterative Error Label Correction}

After one training episode is finished, we can extract deep embeddings of all the samples in the training dataset and perform KNN label correction. Then we move to the next episode with the updated labels. At this point, the neural network may already overfit the noisy labels in the training dataset. As we have mentioned, the quality of embedding extraction by the neural network and the quality of label correction by KNN are correlated. Being trapped in overfitting to noisy labels will adversely affect both qualities.

To overcome overfitting, we regularly reinitialize the neural network model at the beginning of every episode. This will result in a more accurate neural network model in every episode, as long as the label accuracy is improved by label correction. The overall procedure is shown in Algorithm~\ref{Alg:Iter}.

\subsection{Selective Label Correction}

In this paper, we propose two KNN label correction approaches, IterKNN and SelKNN. In IterKNN, the features of all the samples along with their noisy labels are used as reference for KNN label inference as shown in Algorithm~\ref{Alg:InterKNN}. However, datasets for real applications are typically very large. That makes it very expensive to launch KNN classification for every single instance across the whole training dataset. In that sense, a better approach is to selectively choose reference samples for KNN. The most naive selection heuristic is random sampling in the dataset. On the other hand, since the dataset itself is corrupted, the KNN sample set could contain a large portion of samples which have corrupted labels. 

Our objective is to ensure the KNN sample set contains as many clean samples as possible to make it more trustworthy. One simple but effective heuristic to filter clean samples is to utilize the loss information. In general, clean samples are easier to learn and can be learned by the neural network faster. Reversely, noisy labels are more difficult and can only be memorized in the later stage of training~\cite{brodley1999identifying}. 
%The intuition is that clean samples tend to have smaller loss during the training process because it is learned easier and faster, while noisy samples are more likely to have large overall loss it will only be memorized by NN during the later stage of training.
Based on that, we believe cumulative normalized loss is a good factor to determine the likelihood of samples having corrupted labels.

Accordingly, we propose selective label correction algorithm (SelKNN) as shown in Algorithm~\ref{Alg:SelKNN}. We track and accumulate the normalized loss of every image in all previous epochs. At the end of an episode, we rank the cumulative normalized loss for all samples and pick $M$ samples with lowest loss from each class $c$ as the reference of KNN classifier:
\begin{equation}\label{pick_loss}
    B_c ((\bm{x}, \hat{y}), l) = \{(\bm{x}, \hat{y}) \in \hat{S_n} | J(\theta;\bm{x},\hat{y}) \leq l_c, \hat{y} \in c\},
\end{equation}

where $(\bm{\hat{x}}, \hat{y})$ are drawn from the current $\hat{S_n}$, $y$ is the prediction of the current $\mathcal{F}(x, \theta)$, and $l_c$ is the infimum of cumulative loss that class $c$ has $M$ samples with cumulative losses less than or equal to it:
\begin{equation}\label{loss_threshold}
    l_c = \mathrm{inf}\{l: |(\bm{x},\hat{y})| = M,J(\theta;\bm{x},\hat{y}) \leq l\}.
\end{equation}

The labels of those samples in $B_c$ are kept unchanged, while the labels of the remaining samples will be corrected and updated with KNN according to the samples in $B_c$.

\begin{algorithm}
\setlength{\abovecaptionskip}{-0.3cm}
\setlength{\belowcaptionskip}{-0.3cm}
\caption{KNN Label Correction (\textbf{IterKNN})}\label{Alg:InterKNN}
%\hspace*{\algorithmicindent} \textbf{Input:} 
\begin{algorithmic}[1]\\
\textbf{procedure} label\_correction($\hat{D_n}$)\\
\hskip1em $\hat{D_n}' \leftarrow$ \{\}\\
\hskip1em \textbf{for} $(\bm{x}_i, \hat{y_i})$ \textbf{in} $\hat{D_n}$ \textbf{do}\\
\hskip2em Simulate embedding $\bm{e}_i$ with $\mathcal{F}(\bm{x}_i, \theta)$\\
\hskip2em $\hat{y_i}' \leftarrow\ argmax_y \sum_i^n 1\cdot(y=\hat{y_i}, \bm{e}\in N_k(\bm{e}_i))$\\
\hskip2em $\hat{D_n}'\leftarrow \hat{D_n}' \cup (\bm{x}_i, \hat{y_i}')$\\
\hskip1em \textbf{end}\\
\hskip1em \textbf{return} $\hat{D_n}'$\\
\textbf{end procedure}
\end{algorithmic}
%\hspace*{\algorithmicindent} \textbf{Output:} 
\end{algorithm}

\begin{algorithm}
\setlength{\abovecaptionskip}{-0.3cm}
\setlength{\belowcaptionskip}{-0.3cm}
\caption{Selective KNN Label Correction (\textbf{SelKNN})}\label{Alg:SelKNN}
%\hspace*{\algorithmicindent} \textbf{Input:} 
\begin{algorithmic}[1]\\
\textbf{procedure}
selective\_label\_correction($\hat{D_n}$)\\
\hskip1em $B \leftarrow$ \{\}\\
\hskip1em \textbf{for} $c$ \textbf{in} $\mathcal{Y}$ \textbf{do}\\
\hskip2em Find the loss threshold $l_c$ of class $c$ according to Equation \ref{loss_threshold}\\

%\hskip1em $l_c = \mathrm{inf}\{l: |(\bm{x},\hat{y})| = M,J(\theta;\bm{x},\hat{y}) \leq l\}$\\
\hskip2em Select reference samples $B_c$ for KNN according to Equation \ref{pick_loss}\\

%\hskip1em $B_c \leftarrow \{(\bm{x}, \hat{y}) \in \hat{D_n} | J(\theta;\bm{\hat{x}},y) \leq l_c, \hat{y} \in c\}$\\
\hskip2em $B \leftarrow B \cup B_c$\\
\hskip1em \textbf{end}\\
%\hskip1em $\hat{D_n}' \leftarrow$ \{\}\\
\hskip1em $\hat{D_n}' \leftarrow$ $B$\\
\hskip1em \textbf{for} $(\bm{x}_i, \hat{y_i})$ \textbf{in} $\hat{D_n} \setminus B$ \textbf{do}\\
\hskip2em Simulate embedding $\bm{e}_i$ with $\mathcal{F}(\bm{x}_i, \theta)$\\
\hskip2em $\hat{y_i}' \leftarrow\ argmax_y \sum_i^n 1\cdot(y=\hat{y_i}, \bm{e}\in N_{k,B}(\bm{e}_i))$\\
\hskip2em $\hat{D_n}'\leftarrow \hat{D_n}' \cup (\bm{x}_i, \hat{y_i}')$\\
\hskip1em \textbf{end}\\
\hskip1em \textbf{return} $\hat{D_n}'$\\
\textbf{end procedure}
\end{algorithmic}
%\hspace*{\algorithmicindent} \textbf{Output:} 
\end{algorithm}

\subsection{Design of Loss Function}

The DNN part in the framework keeps being updated with regard to the loss function. If only the corrected labels are used in training, the network will reach self-convergence quickly. It is because of the nature of KNN: KNN will correct the noisy or hard samples, and result in a dataset that is easier for the network to learn. To resolve this quick self-convergence problem, we always keep a copy of the original noisy dataset along with the corrected dataset, which is updated after every episode. We implement a total loss that is a convex combination of the losses on the two datasets:
\begin{equation}
    J_{train} = (1-\gamma)\cdot J(\mathcal{F}(\bm{x}, \theta), \hat{y}) +  \gamma \cdot J(\mathcal{F}(\bm{x}, \theta), \tilde{y}),
\end{equation}

where $\gamma$ is the weight coefficient for the total loss $J_{train}$, which is distributed between the original noisy dataset $\tilde{D_n}$ and the current KNN corrected dataset $\hat{D_n}$.
%with the original noisy label $\tilde{y}$ as target. 
Initially $\gamma$ is set as $1.0$, meaning in the first iteration we only consider the original noisy labels and ignore the KNN predicted labels. This makes sense since the DNN classifier has not been trained and no meaningful features can be obtained from it. To improve convergence of the algorithm, we decay $\gamma$ by $\sigma$ for every training iteration after the current labels are updated. As the episode index grows, our training loss tends to focus on the KNN predicted labels. Eventually, when $\gamma$ gets close to 0, the neural network distill itself and the algorithm will converge. We realize our final loss is similar to bootstrapping loss~\cite{reed2014training}. The difference is that bootstrapping uses prediction from classifier itself while we use prediction from KNN. We will empirically show KNN prediction is more robust in Sec.~\ref{Sec: KNN robustness}. Overall architecture is illustrated in Figure~\ref{fig:short}.

\section{Experiments and Evaluation}

%%%%%%%%%%%%%%%%%%%%%%%%%%%%%%%%%%%%%%%%%%%%%%%%%%%%%%%%%%%%%%%%%%%%%%%%%%%
%A figure showing KNN and CNN prediction accuracy. Prediction from deep KNN is more robust against noise than prediction from CNN. This is because the even though the prediction result of CNN is largely impacted by label noise, the noise will impact the last few layers more than the previous layers. That means CNN can still capture useful features during the training with the label corruption. 4 curves:
%1 CNN accuracy on training dataset, 
%2 KNN accuracy if all ground truth is given
%3 KNN accuracy if only correct labels ground truth is given, use ground truth to KNN databse only
%4 KNN accuracy using our algorithm to filter and predict
%5 KNN accuracy using whole real dataset

%show the power of KNN, if we know some portion of data is clean. Then training on the whole dataset, but use clean data as KNN database to classify the labels for whole dataset. The result is much better than CNN prediction. Using the aforementioned curve to prove

%%%%%%%%%%%%%%%%%%%%%%%%%%%%%%%%%%%%%%%%%%%%%%%%%%%%%%%%%%%%%%%%%%%%%%%%%%%%%%%
% a figure to show the impact of K
%multiple different curve

% a figure to show 3 layers
%% probably in supplementry work: a figure to show L2+majority, L2+weighted average, Cos+majority, Cos+weighted average,

%probably another figure to show each iteration there is updating.
%%%%%%%%%%%%%%%%%%%%%%%%%%%%%%%%%%%%%%%%%%%%%%%%%%%%%%%%%%%%%%%%%%%%%%%%%%%%

%performance of our approach depends on the clean/noisy label detection accuracy.

We conduct experiments and evaluate our approach on 3 popular synthetic datasets (MNIST, CIFAR-10, CIFAR-100), and 1 large-scale real life dataset (Clothing1M). All experiments are implemented using the PyTorch framework.

\subsection{Noise Settings}

We consider both symmetric and asymmetric label noises in our experiments.

\textbf{Symmetric Noise:} %With symmetric label noise, labels of every class have equal probabilities to be flipped to any other class. 
Given the noise level $\pi$, the label of every sample has a probability $\pi$ to be flipped to another class uniformly at random:
\begin{equation}
%\tilde{y}=
\left\{
\begin{aligned}
Pr(\tilde{y_i}=y_i) &= 1-\pi, \\
\forall j \neq i : Pr(\tilde{y_i}=y_j) &= \frac{\pi}{|\mathcal{Y}|-1}. \\
\end{aligned}
(i, j \in \mathcal{Y})\right.
\end{equation}

\textbf{Asymmetric Noise:} We use a similar configuration as discussed in~\cite{patrini2017making}. We inject asymmetric label noises to mimic the part of the mistakes made between similar classes. For MNIST, we use the following class transitions: 7$\rightarrow$1, 2$\rightarrow$7, 5$\leftrightarrow$6, and 3$\rightarrow$8. For CIFAR-10, we have \textit{truck}$\rightarrow$\textit{automobile}, \textit{bird}$\rightarrow$\textit{airplane}, \textit{cat}$\leftrightarrow$\textit{dog}, and \textit{deer}$\rightarrow$\textit{horse}. For CIFAR-100, we pair classes in a similar fashion.

\begin{figure*}
    \centering
    \begin{subfigure}[b]{0.325\textwidth}
        \includegraphics[width=\textwidth]{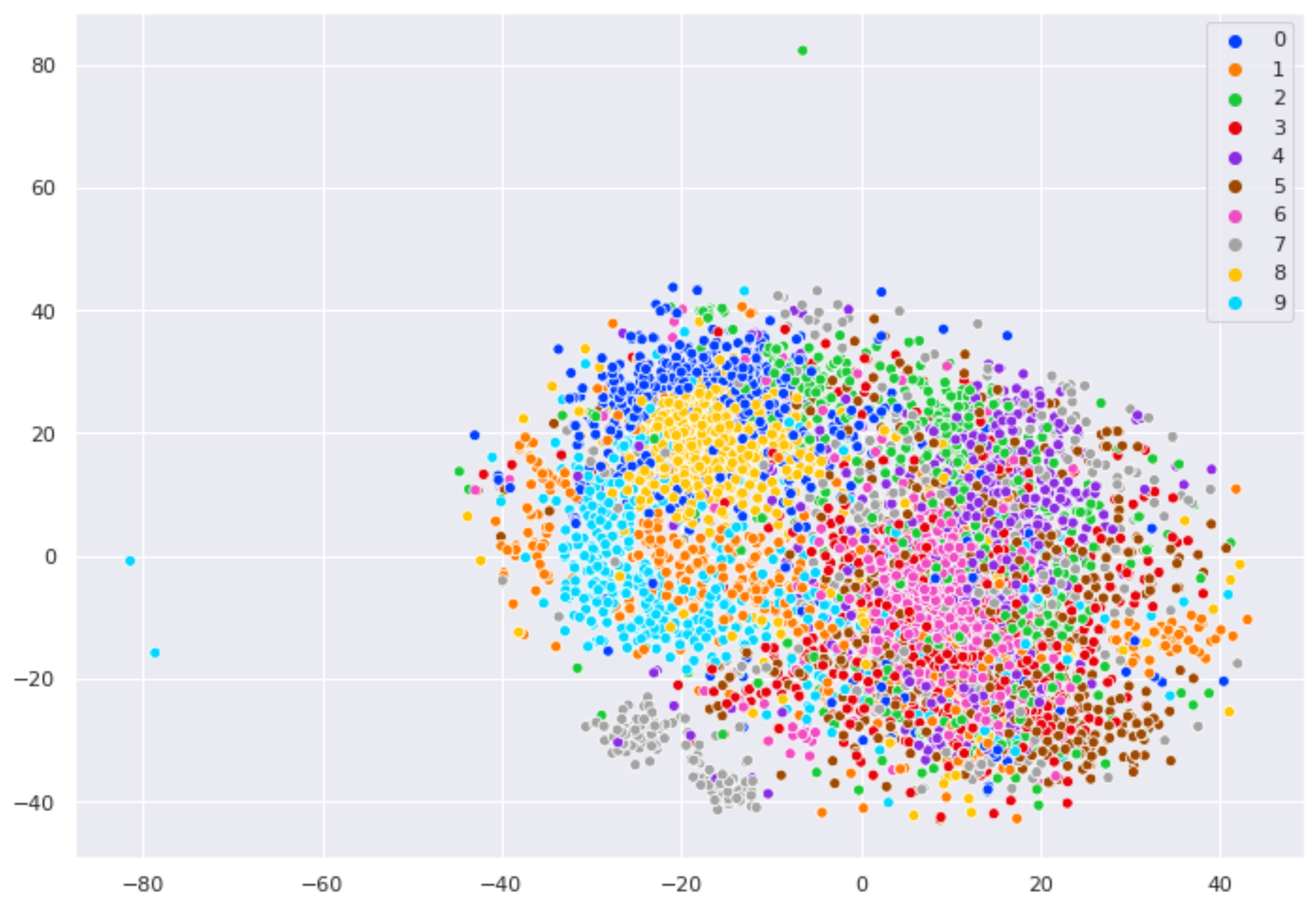}
        \caption{First Convolutional Layer Output}
        \label{fig:gull}
    \end{subfigure}
    \begin{subfigure}[b]{0.325\textwidth}
        \includegraphics[width=\textwidth]{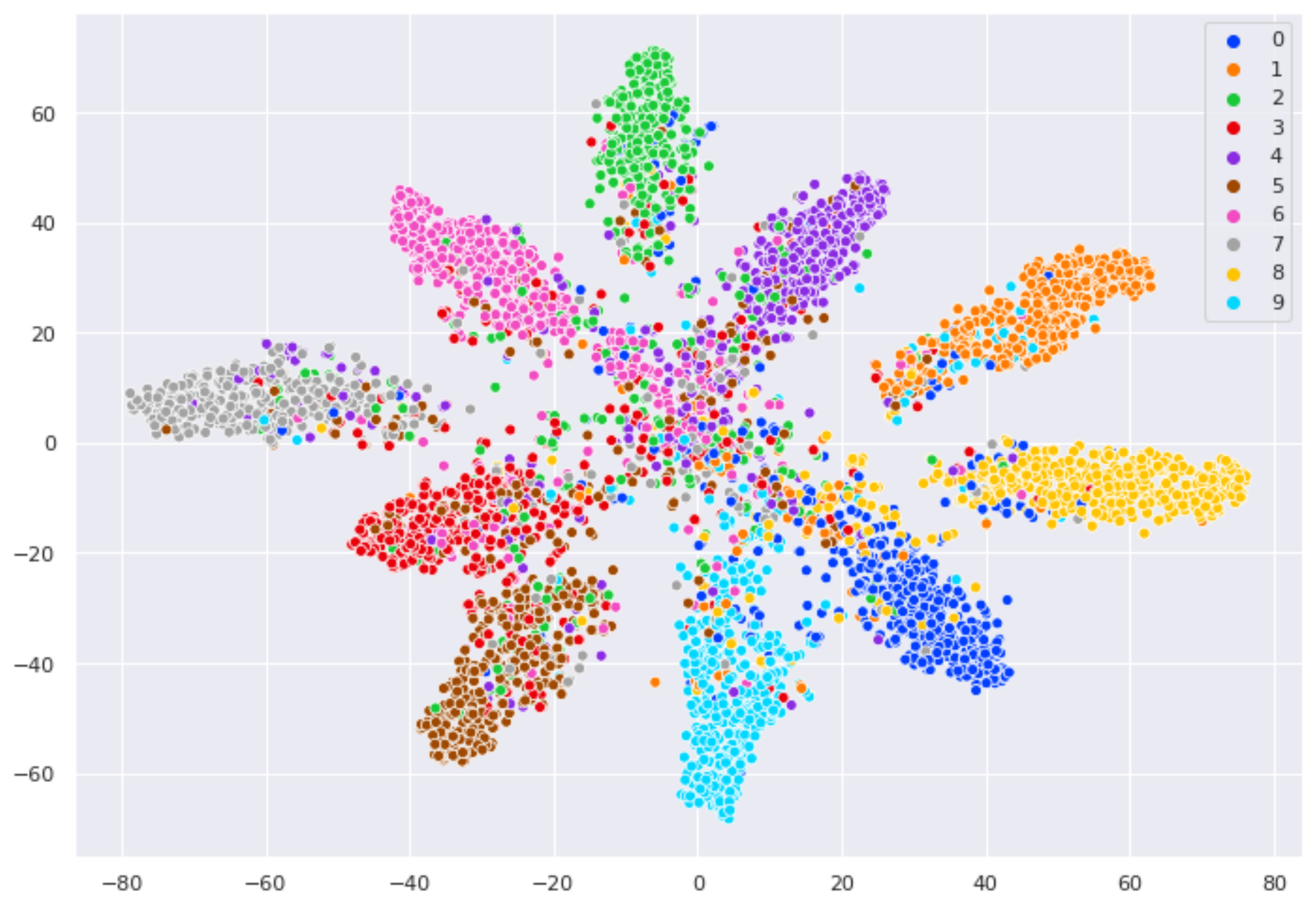}
        \caption{Last Convolutional Layer Output}
        \label{fig:tiger}
    \end{subfigure}
    \begin{subfigure}[b]{0.325\textwidth}
        \includegraphics[width=\textwidth]{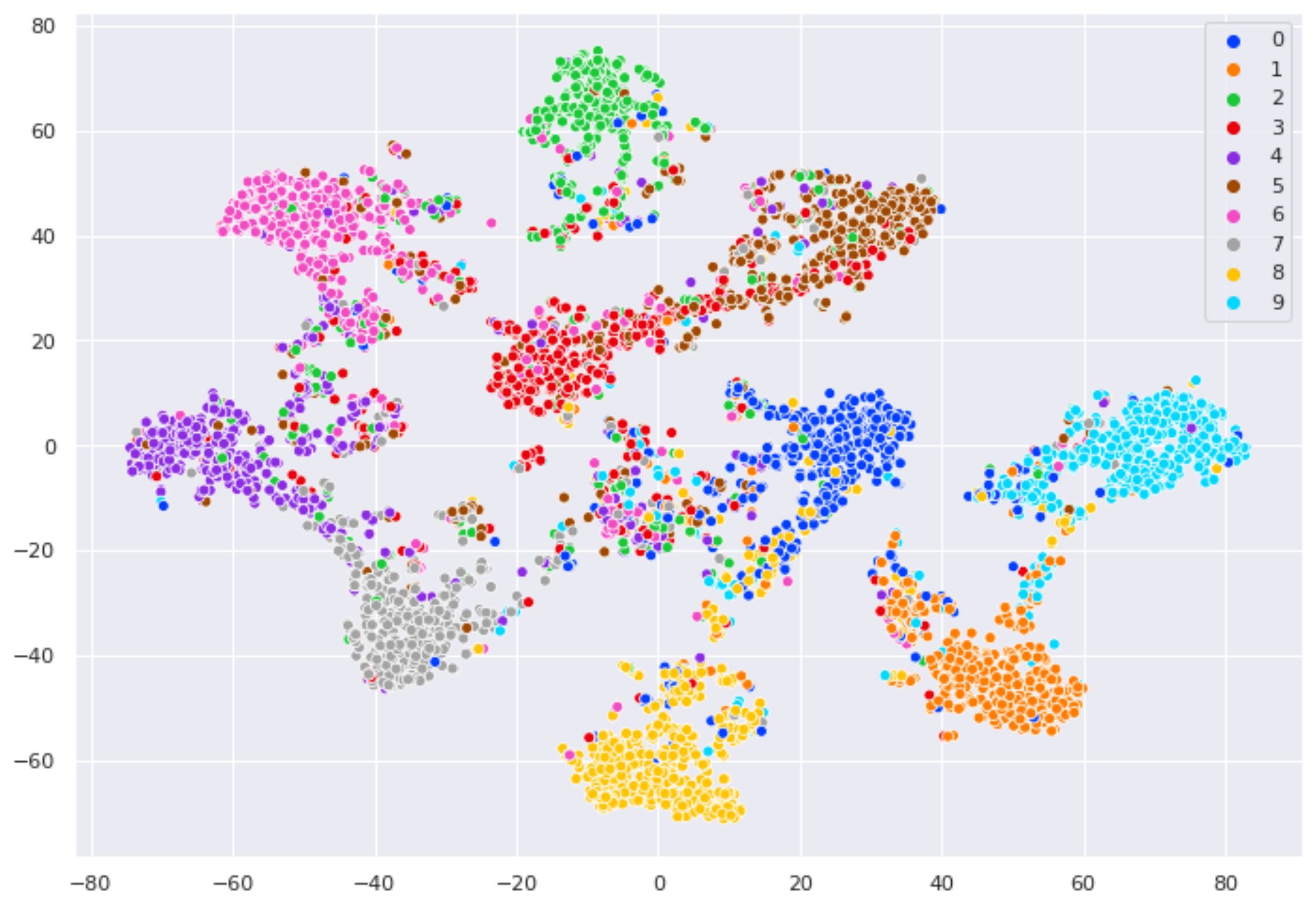}
        \caption{Fully Connected Layer Output}
        \label{fig:mouse}
    \end{subfigure}
    \caption{t-SNE 2D embeddings of outputs from different layers after 1 training episode. (CIFAR-10 with $60\%$ symmetric noisy labels)}\label{fig:deep representation}
\end{figure*}

\subsection{Hyper-parameter Settings}

We use Adam optimizer with weight decay of 1e-4 and initial learning rate of 0.001 for all experiments.

\textbf{MNIST:} We use a 5-layer CNN model with 4 convolution layers followed with a fully connected layer. Each training episode has 40 epochs and the learning rate is decreases by 10 times on epoch 20 and epoch 30. The batch size is 256.

\textbf{CIFAR-10:} We use Pre-act Resnet 32~\cite{he2016identity}. Each episode of training has 120 epochs and the learning rate is decreased by 10 times on epoch 60 and epoch 90. For data augmentation, we use random crop, random horizontal flip, random affine and color jetter. The training batch size is 512.

\textbf{CIFAR-100:} We use Pre-act Resnet 56. Each episode of training has 120 epochs and the learning rate is decreased by 10 times on epoch 80 and epoch 120. Data augmentation and training batch size are same to those of CIFAR-10.

%%%%%%%%%%%%%%%%%%%%%%%%%%%%%%%%%%%%%%%%%%%%%%%%%%%%%%%%%%%%%%%%%%%%%%%%%%%%%%%%%%%%%%%%%%%%%%
\begin{table*}[h!]
%\scriptsize
\setlength{\abovecaptionskip}{-0.2cm}
\setlength{\belowcaptionskip}{-0.2cm}
\begin{center}
\begin{tabular}{c|cccc|ccc}
\hline
 methods & \multicolumn{4}{c|}{symmetric noise rate} & \multicolumn{3}{c}{asymmetric noise rate}  \\\cline{2-8}
 & 20\% & 40\% & 60\% & 80\%  & 20\% & 30\% & 40\%   \\
\hline
\multicolumn{8}{c}{MNIST} \\
\hline 
 CE  & 89.91$\pm$0.02 & 68.75$\pm$0.04 & 44.35$\pm$0.24 & 24.77$\pm$0.37  & 94.22$\pm$0.07 & 86.40$\pm$0.12 & 80.33$\pm$0.09 \\
 %Forward  & 92.44$\pm$0.19 &  80.67$\pm$0.46 & 61.45$\pm$0.43 & 29.34$\pm$0.51  & 96,88$\pm$0.22 & 91.66$\pm$0.31 & 87.33$\pm$0.37 \\
 %Bootstrapping  & 87.39$\pm$0.09 &  71.35$\pm$0.24 & 50.23$\pm$0.08 & 26.33$\pm$0.31 & 93.77$\pm$0.07 & 89.87$\pm$0.12 & 81.26$\pm$0.19 \\
 Joint Opt~\cite{Tanaka_2018_CVPR}  & 94.89$\pm$0.07 &  93.94$\pm$0.16 & 91.03$\pm$0.46 & 66.55$\pm$1.05  & 94.79$\pm$0.26 & 93.01$\pm$0.24 & 91.78$\pm$0.33 \\
 PENCIL~\cite{Yi_2019_CVPR}  &  95.34$\pm$0.21 &  93.89$\pm$0.72 & 93.06$\pm$0.53 & 70.25$\pm$0.86 & 96.22$\pm$0.18  & 94.38$\pm$0.21 & 90.15$\pm$0.42  \\
  SL~\cite{Wang_2019_ICCV} &  \textbf{98.77$\pm$0.07} & 97.65$\pm$0.19 & 95.27$\pm$0.35 & 68.04$\pm$0.33  & 98.61$\pm$0.21 & 97.11$\pm$0.81 & 96.15$\pm$0.69 \\
  \textbf{IterKNN} & 98.33$\pm$0.27 & 96.35$\pm$0.22 & 94.33$\pm$0.32 & 68.81$\pm$0.49 & 98.65$\pm$0.11 & 97.27$\pm$0.15 & 94.18$\pm$0.41 \\
  \textbf{SelKNN}  & 98.25$\pm$0.12 & \textbf{97.79$\pm$0.19} & \textbf{96.12$\pm$0.31} & \textbf{70.88$\pm$0.29} & \textbf{98.69$\pm$0.08} & \textbf{97.80$\pm$0.17} & \textbf{96.39$\pm$0.18} \\

\hline
\multicolumn{8}{c}{CIFAR10} \\
\hline 
 CE & 83.04$\pm$0.18 & 66.75$\pm$0.11 & 46.98$\pm$0.21 & 21.33$\pm$0.36  & 86.16$\pm$0.24 & 80.91$\pm$0.13 & 70.51$\pm$0.76 \\
 %Forward & 84.14$\pm$0.44 &  71.44$\pm$0.74 & 52.13$\pm$0.83 & 42.18$\pm$0.96 & 87.88$\pm$0.21 & 83.69$\pm$0.57 & 75.56$\pm$0.52  \\
 %Bootstrapping   & 83.16$\pm$0.19 &  69.18$\pm$0.41 & 59.23$\pm$0.39 & 29.75$\pm$0.95  & 87.67$\pm$0.09 & 81.11$\pm$0.26 & 73.06$\pm$0.47 \\
 Joint Opt~\cite{Tanaka_2018_CVPR}  & 91.61$\pm$0.74 &  87.75$\pm$0.36 & 84.02$\pm$0.42 & 58.46$\pm$0.97  & 91.16$\pm$0.09 & 89.41$\pm$0.21 & 86.76$\pm$0.81 \\
 PENCIL~\cite{Yi_2019_CVPR}  & \textbf{92.45$\pm$0.81} &  88.32$\pm$0.49 & 84.42$\pm$0.98 & 59.02$\pm$0.72  & 90.21$\pm$0.16 & 89.36$\pm$0.32 & 87.59$\pm$0.45 \\
  SL~\cite{Wang_2019_ICCV}   & 87.89$\pm$0.07  &  84.36$\pm$0.09 & 79.75$\pm$0.18 & 55.36$\pm$0.42  & 87.44$\pm$0.22 & 85.24$\pm$0.25 & 80.32$\pm$0.29 \\
  \textbf{IterKNN}  & 89.43$\pm$0.14 & 87.34$\pm$0.23 & 80.19$\pm$0.21 & 56.84$\pm$0.38   & 90.29$\pm$0.22 & 89.28$\pm$0.18 & 87.15$\pm$0.33 \\
  \textbf{SelKNN}   & 91.99$\pm$0.44 & \textbf{89.25$\pm$0.39} & \textbf{85.65$\pm$0.92} & \textbf{59.65$\pm$0.27} & \textbf{91.38$\pm$0.25} & \textbf{89.75$\pm$0.32} & \textbf{88.06$\pm$0.28}\\

\hline
\multicolumn{8}{c}{CIFAR100} \\
\hline 
 CE  & 60.13$\pm$0.20 & 51.25$\pm$0.52 & 36.75$\pm$0.81 & 19.35$\pm$0.79  & 61.16$\pm$0.74 & 56.27$\pm$0.78 & 47.35$\pm$0.78 \\
 %Forward  & 63,25$\pm$0.55 &  53.61$\pm$0.87 & 29.25$\pm$0.48  & 21.35$\pm$1.06  & 62.11$\pm$0.29 & 58.79$\pm$0.72 & 50.92$\pm$0.83 \\
 %Bootstrapping   & 61.89$\pm$0.32 &  51.80$\pm$0.30 & 31.14$\pm$0.89 & 17.40$\pm$1.21  & 62.04$\pm$0.31 & 57.66$\pm$0.62 & 51.47$\pm$0.72 \\
 Joint Opt~\cite{Tanaka_2018_CVPR}   & 68.06$\pm$0.88 & 64.79$\pm$0.91 & 54.29$\pm$0.97 & 27.78$\pm$0.94  & 69.13$\pm$0.39 & 68.49$\pm$0.36 & 59.23$\pm$0.70 \\
 PENCIL~\cite{Yi_2019_CVPR}  & 71.14$\pm$0.34 &  68.75$\pm$0.62 & 56.02$\pm$0.66 & 26.33$\pm$0.58  & \textbf{73.15$\pm$0.85} & 71.02$\pm$0.63 & 61.48$\pm$0.94 \\
  SL~\cite{Wang_2019_ICCV}   & 64.85$\pm$0.29 & 59.42$\pm$0.34 & 45.85$\pm$0.46 & 22.81$\pm$0.79  & 65.19$\pm$0.21 & 63.37$\pm$0.45 & 60.01$\pm$0.65  \\
 \textbf{IterKNN}   & 69.15$\pm$0.30 & 62.48$\pm$0.91 & 53.36$\pm$0.47 & 28.93$\pm$0.98  & 69.92$\pm$0.42 & 68.13$\pm$0.37 & 62.14$\pm$0.39 \\
  \textbf{SelKNN}  & \textbf{72.75$\pm$0.68} & \textbf{69.88$\pm$0.64} & \textbf{57.43$\pm$1.23} & \textbf{32.91$\pm$0.57}  & 72.03$\pm$0.26 & \textbf{71.09$\pm$0.45} & \textbf{63.32$\pm$1.39} \\

\hline
\end{tabular}
\end{center}
\caption{Accuracy(\%) from the last training epoch on 3 datasets under different noise settings. We run 10 trials, report mean and std. Our approach achieves best results under most settings.}\label{table:baseline_comparison}
\end{table*}

\begin{table}[h!]
%\scriptsize
\setlength{\abovecaptionskip}{-0.2cm}
\setlength{\belowcaptionskip}{-0.2cm}
\begin{center}
\begin{tabular}{c|ccc}
\hline
 methods & \multicolumn{3}{c}{symmetric noise} \\
  & 20\% & 40\% & 80\% \\
\hline\hline 
 Joint Opt& 93.35$\pm$0.32 & 90.31$\pm$0.36 &  58.72$\pm$0.85  \\
PENCIL & 94.36$\pm$0.36 & 91.21$\pm$0.52 &  59.14$\pm$0.16  \\
IterKNN & \textbf{94.47$\pm$0.25} & 89.16$\pm$0.28 &  56.95$\pm$0.79  \\
SelKNN & 93.29$\pm$0.25 & \textbf{91.32$\pm$0.22} &  \textbf{59.94$\pm$0.52} \\
\hline
\end{tabular}
\end{center}
\caption{Label recovery rate(\%) on CIFAR-10 with different symmetric noise levels. We run 10 trials, report mean and std.}\label{table:label_recovery_sym}
\end{table}

We compare our approach with the follow baselines:

\textbf{Training with cross entropy:} This is the most basic training setting, in which the neural network is trained directly with noisy labels using cross entropy loss.
%\textbf{Training with Bootstrapping~\cite{reed2014training}: }
%This approach conducts training with new labels generated by a convex combination of the raw labels and the predicted labels. We report results on soft bootstrapping, where the soft predicted labels are used.
%\textbf{Forward~\cite{patrini2017making}:}
%This work proposes a loss correction approach using the noisy transition matrix $T$. However, for fair comparison we assume the $T$ is not known and needs to be estimated. In our experiment, we estimate $T$ based on the softmax output of the DNN trained from noisy labels.
\textbf{Symmetric Loss~\cite{Wang_2019_ICCV}:}
This approach uses a combination of cross entropy loss and reverse cross entropy loss. In that way it encourages learning hard samples without overfitting noisy labels.
\textbf{Joint Optimization~\cite{Tanaka_2018_CVPR}:}
This method optimizes the loss by updating network parameters and class labels alternatively.
\textbf{PENCIL~\cite{Yi_2019_CVPR}:}
This approach is similar to Joint Optimization. The only major difference is that differentiable psuedo labels are created and updated together with model parameters simultaneously. Both Joint Optimization and PENCIL add two extra losses, entropy loss and regularization loss, which require prior knowledge of the true class distribution.

All the above baselines are re-implemented according to their open-source codes with minor modifications to fit our settings. We always set \textit{k} value to $100$, using L2 distance metric, and use majority voting for label inference on both IterKNN and SelKNN. We perform in total 10 episodes of training together with label updating for every time of execution. Symmetric cross entropy loss is adopted to encourage learning hard examples. Weight coefficient $\gamma$ is initially $1.0$ and decayed by $1.2$ every episode. For SelKNN, we pick up top $M\%$ of images from each class as KNN reference samples in order to update the labels of the the remaining $1-M\%$ samples. $M$ is set to $20$ for first episode and is incremented by $10$ every episode afterwards, until it reaches $100$. We can generally increase $M$ because more and more corrupted labels are cleaned up. When $M$ reaches $100$, SelKNN will reduce to IterKNN to facilitate further label correction. The comparison results are shown in Table~\ref{table:baseline_comparison}. SelKNN achieves state-of-the-art performance on most noise settings, especially when noise rate is high while IterKNN has competitive performance when the noise rate is low. Another observation is the variance of the accuracy among multiple trials is low when the noise rate is low due to less randomness. When the noise rate increases, the variance of test accuracy obtained from SelKNN is least affected since it only pick highly possible clean samples to update labels. This result further validates the robustness of our SelKNN approach.

\begin{figure}[h!]
\begin{center}
%\fbox{\rule{0pt}{2in} \rule{.9\linewidth}{0pt}}
\includegraphics[width=0.49\textwidth]{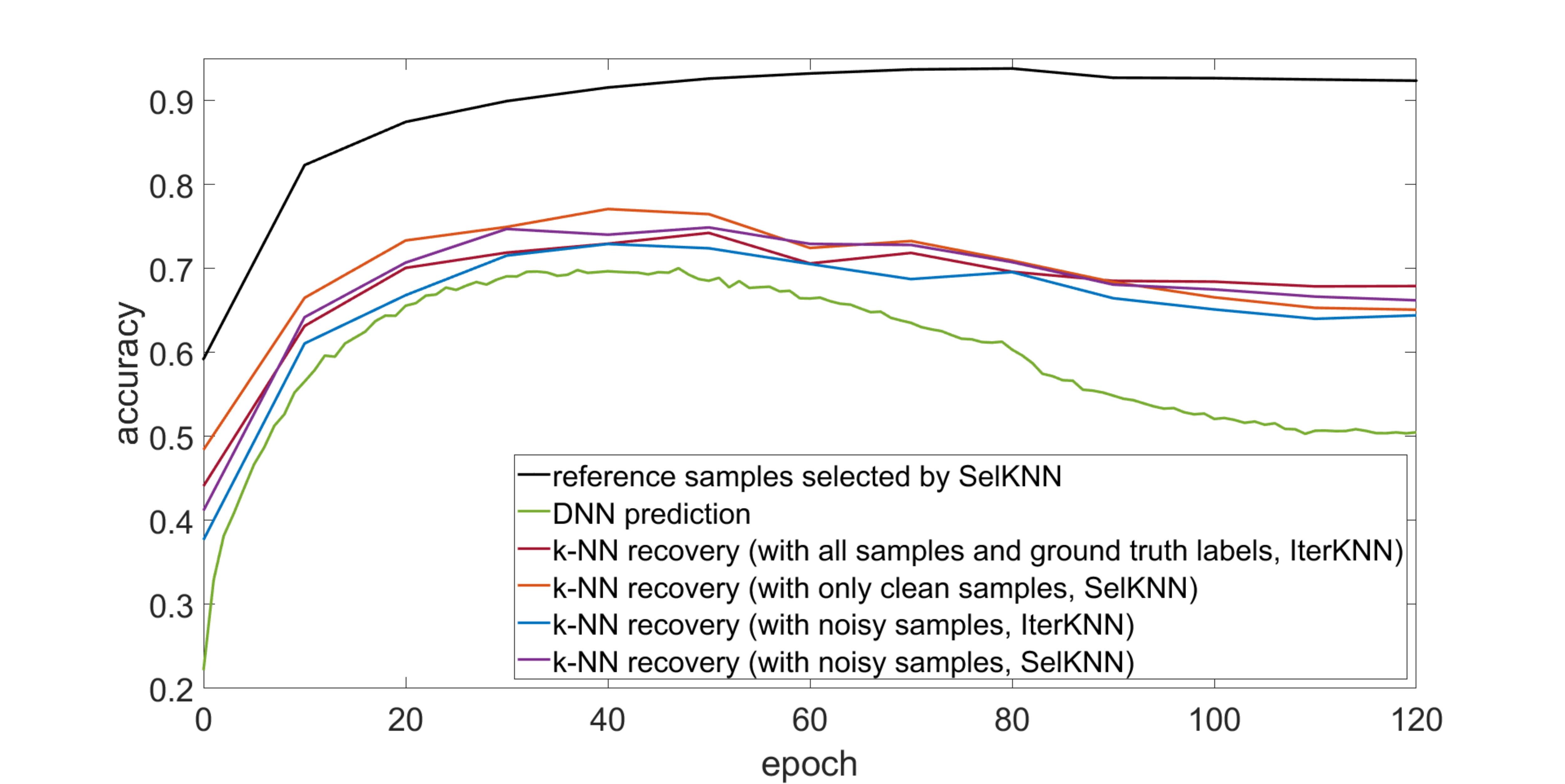}
\end{center}
    \caption{Ground truth label inference accuracy in the first episode with different prediction methods, including DNN prediction, IterKNN prediction using noisy labels or ground-truth labels, SelKNN prediction using filtered labels or known clean labels ($20\%$ of whole dataset). The top curve is the clean rate of reference samples, which converges to around $92\%$. (CIFAR-10 with $60\%$ symmetric noisy labels)}
\label{fig:robustness}
\end{figure} 
\subsection{Robustness and Effectiveness of KNN}\label{Sec: KNN robustness}
To demonstrate the robustness of KNN over DNN, we perform 5 types of 1-episode training on CIFAR-10 with $60\%$ symmetric noise. Here we employ cross entropy loss to better differentiate the robustnesses of different predictions since it is relatively weak against label noise. We keep track of the ground-truth label inference accuracy as training proceeds. The results and details are illustrated in Fig.~\ref{fig:robustness}. All types of predictions overfit the data noise to some extent at the later stage of the episode.
We let the DNN trained on noisy dataset to extract embeddings as well as predict labels for input images. When IterKNN and SelKNN are allowed to run classification on clean labels, their predictions are less subject to over-fitting than DNN prediction. 
%When classifier is trained on corrupted labels but suppose IterKNN and SelKNN know the ground-truth labels of the found neighbors for every sample, their predictions are less subject to over-fitting than DNN prediction. 
This shows the DNN can still learn from noisy datasets, so that it derives similar features for similar images, which validates the robustness of KNN based approach. On the other hand, when no ground-truth or clean labels are provided, both of IterKNN and SelKNN have slightly worse performance, but are still clearly better than the DNN prediction. All those observations lead to the following two conclusions: (1) our IterKNN and SelKNN should be more effective than those self-learning approaches using DNN prediction to recover labels~\cite{Tanaka_2018_CVPR}.  (2) SelKNN can be improved with a better filtering method (\textit{e.g.},~\cite{Huang_2019_ICCV}) to select clean reference samples.
%To demonstrate the robustness of KNN over CNN, we perform 5 types of 1-episode training on CIFAR-10 with $60\%$ symmetric noise. Here we employ cross entropy loss to better differentiate the robustnesses of different predictions since it is relatively weak against label noise. We keep track of the ground-truth label inference accuracy as training proceeds. The results and details are illustrated in Fig.~\ref{fig:robustness}. All types of predictions overfit the data noise to some extent at the later stage of the episode. When IterKNN is performed with all ground-truth labels or SelKNN is performed with known clean labels as reference, the over-fitting is less severe than that of the CNN prediction. This shows the deep neural network can still learn on noisy datasets, in a sense that it derives similar features for similar images, which validates the robustness of KNN based approach. On the other hand, when no ground-truth or clean labels are provided, both of IterKNN and SelKNN have slightly worse performance, but are still clearly better than the CNN prediction. This observation leads to the following two conclusions: (1) our SelKNN should be more effective at label correction than those self-learning approaches using CNN prediction to update labels~\cite{Tanaka_2018_CVPR}.  (2) SelKNN can be improved with a better filtering method (\textit{e.g.},~\cite{Huang_2019_ICCV}) to select clean reference samples.

\begin{figure}%[h]
\begin{center}
%\fbox{\rule{0pt}{2in} \rule{.9\linewidth}{0pt}}
\includegraphics[width=0.49\textwidth]{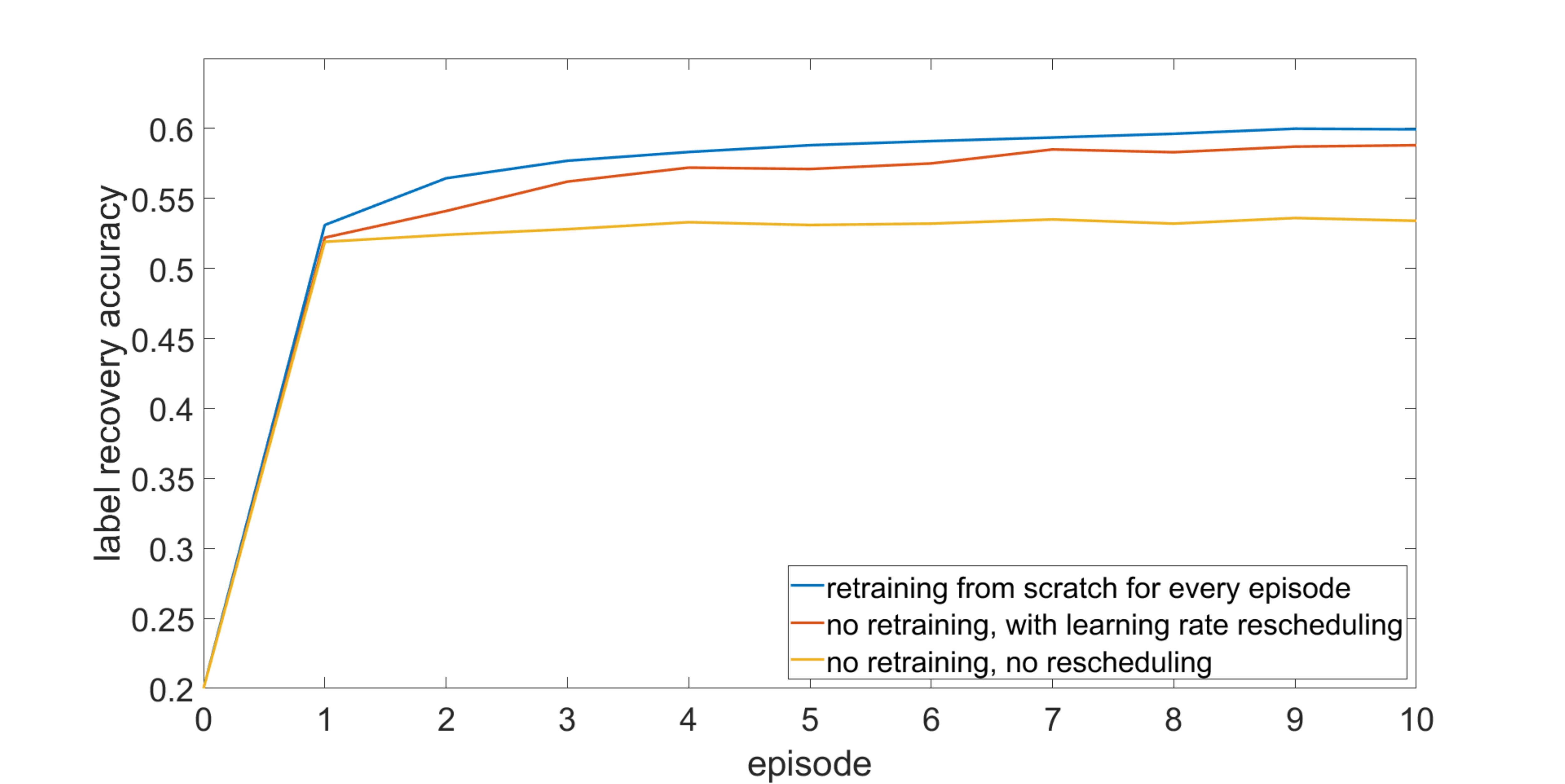}
\end{center}
    \caption{Label recovery rate for SelKNN under different training settings versus number of episodes. 
    ``No retraining'' means training continues on the current neural network instead of re-initialized from scratch for every next episode. (CIFAR-10 with $80\%$ symmetric noisy labels)}
\label{fig:iter_training}
\end{figure} 

\subsection{Iterative Retraining}

Iterative retraining can correct more noisy labels and further improve the performance of the classifier, especially when noise level is extremely high. As shown in Fig.~\ref{fig:iter_training}, the main proportion of label recovery is achieved in the first three episodes, yet some incremental recovery happens in the remaining episodes. The main purpose of retraining from scratch is to escape from over-fitting to the current noisy labels. Another approach to overcome over-fitting is to continue training but periodically reschedule the learning rate. When labels have been updated, the learning rate is also reset to a large value so that the model can transit to an under-fitting state. As shown in Fig.~\ref{fig:iter_training}, performance cannot be further improved for later episodes if training continues without resetting the learning rate. It is because with small learning rate the model cannot escape from over-fitting noisy labels. We also found out retraining from re-initialized neural network performs slight better than simply rescheduling learning rate since re-initialization makes neural network completely forget noisy labels. We compare the final recovery accuracy with baselines in Table~\ref{table:label_recovery_sym}. IterKNN performs well when noise level is low while SelKNN beats other methods on highly noisy dataset.

%In our proposed algorithm, we train the reinitialized model after labels are updated. We do not continue training the model with the scheduled learning rate because it already overfits label noise in the dataset. Besides retraining, resetting the learning rate to a large value is another way to escape from overfitting. A natural solution is to cyclically adjusting learning rate after each label correction. We compare this method with iterative retraining.

%Compare with retraining and without retraining(continue training), using a figure, may run 3 iteration of retraining, total epochs same, figure should be on the high noise. Maybe also discuss and compare cyclic learning(learning rate)

%high noise needs more training iterations

%As iteration goes on, the training epochs between two consecutive label correction round increases, the stage I epochs also increases. High noise level needs a smaller number of stage I training to avoid overfitting bad samples.
%Maybe use an expression to formulate it

%iterative KNN label correction and retraining is more helpful for training dataset with high noise levels.

\begin{figure}[h!]
\begin{center}
%\fbox{\rule{0pt}{2in} \rule{.9\linewidth}{0pt}}
\includegraphics[width=0.49\textwidth]{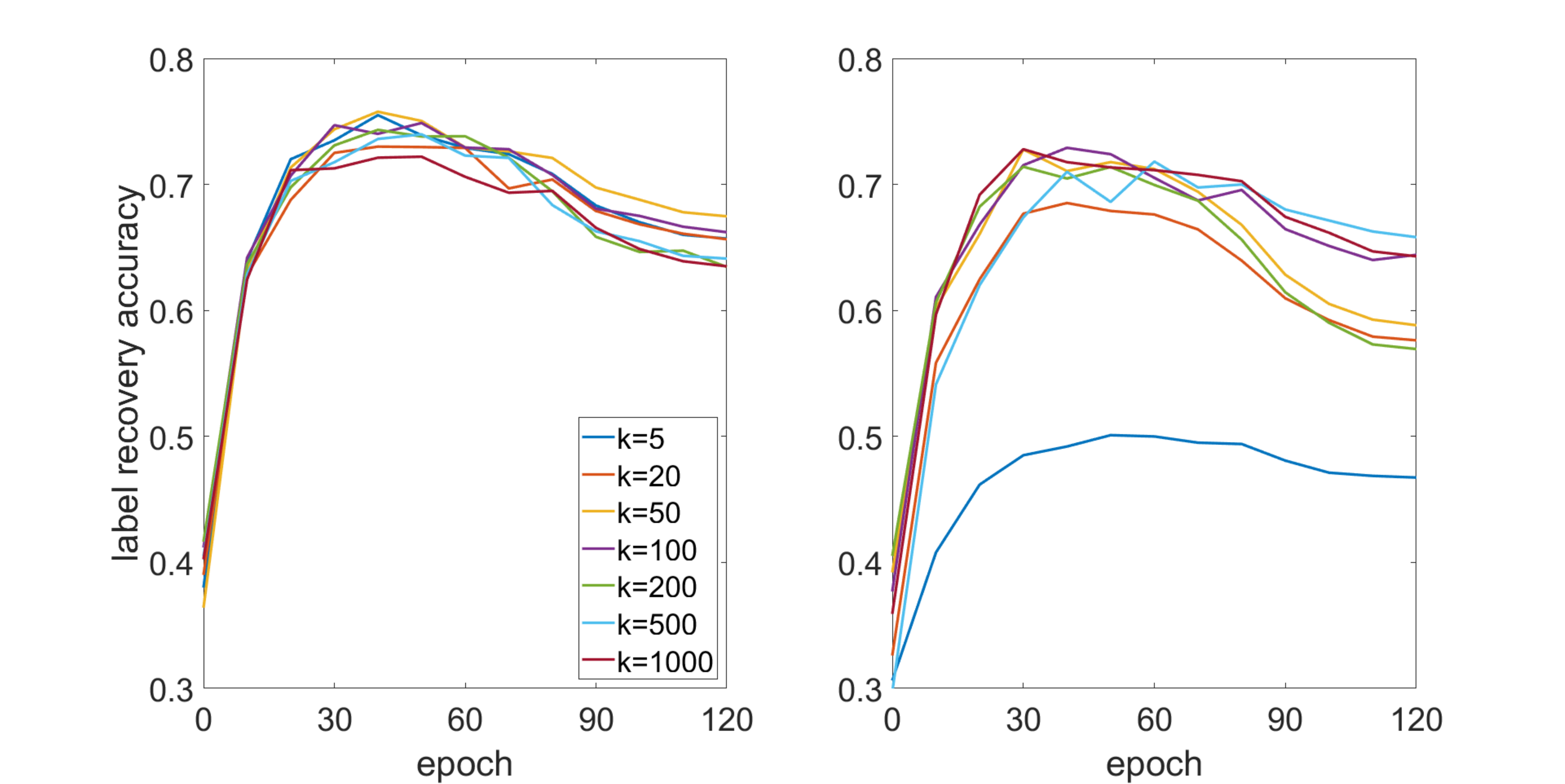}
\end{center}
    \caption{Ground-truth label inference accuracy of SelKNN (left) and IterKNN (right) with different \textit{k} in the first episode. (CIFAR-10 with $60\%$ symmetric noisy labels)}
\label{fig:k_rev}
\end{figure}

\subsection{Impact of \textit{k} Value}
%Noise level high, small decay, noise level low, large decay of alpha and smaller number of retraining iterations
Since $k$ value can have a large impact on the performance of neighbor classification, we also investigate how IterKNN and SelKNN can be affected by $k$. We select 7 different \textit{k} values ($5,20,50,100,200,500,1000$) and track the label prediction accuracy within one training episode without any label correction. The prediction is made after every 10 epoch and is shown in Fig.~\ref{fig:k_rev}. As can be seen, SelKNN is relatively invariant to $k$ and all the choices of $k$ consistently outperform the label prediction from classifier itself during the episode. It is also clear that IterKNN is more sensitive to $k$ and requires a reasonably large $k$ value to obtain good performances. This is because for IterKNN, the whole noisy training dataset is used as reference for KNN prediction. When $k$ is too small, it is more likely all nearest neighbors are corrupted. So even if all the neighbors as well as the queried image belong to the same ground-truth class, KNN will still infer an incorrect labels.

\begin{figure}[h!]
\begin{center}
%\fbox{\rule{0pt}{2in} \rule{.9\linewidth}{0pt}}
\includegraphics[width=0.49\textwidth]{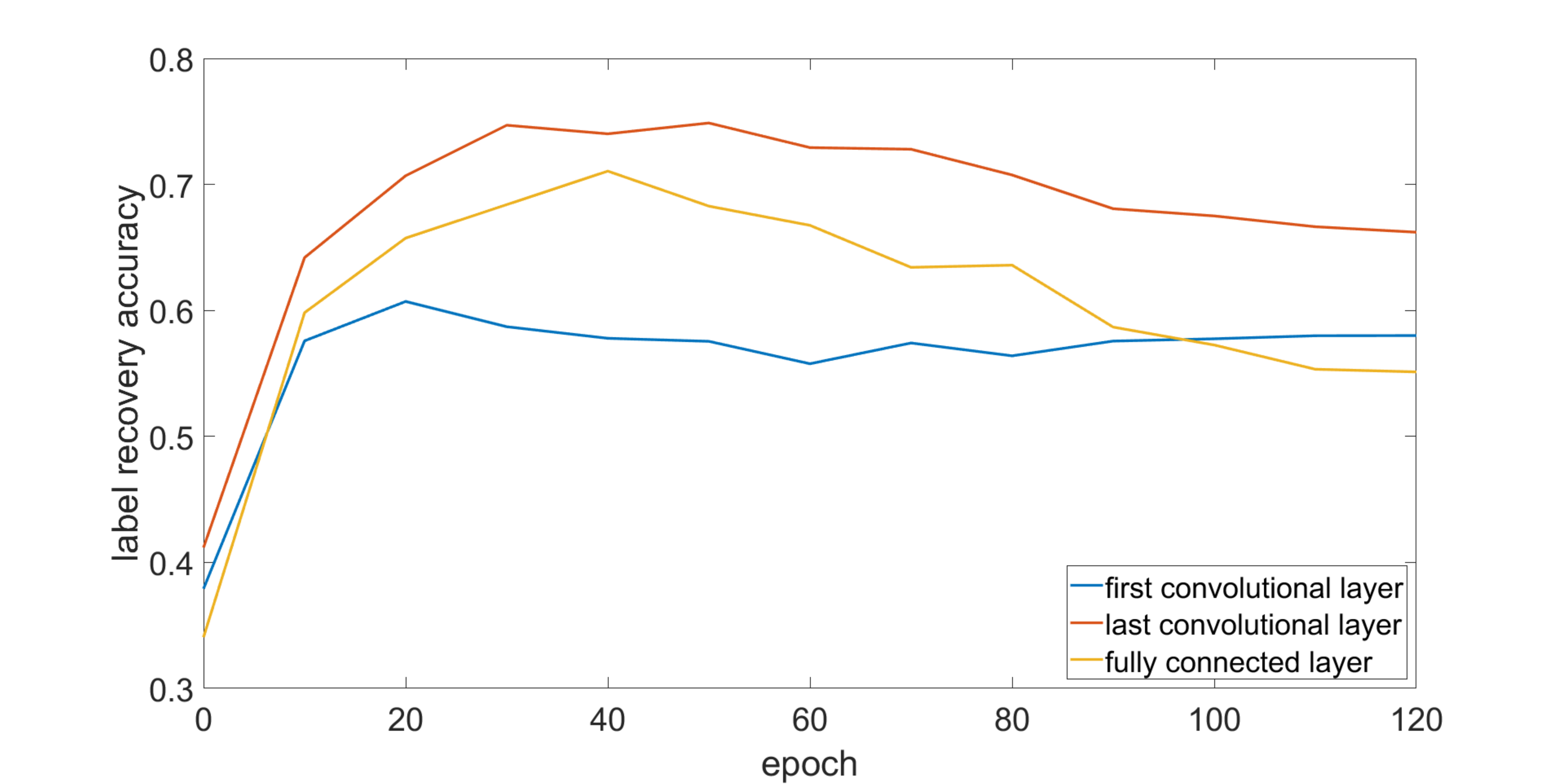}
\end{center}
    \caption{Ground-truth label inference accuracy of SelKNN with different layer features in the first episode. (CIFAR-10 with $60\%$ symmetric noisy labels)}
\label{fig:deep_feature}
\end{figure} 

\subsection{Choice of Deep Features}
A fundamental question for KNN-based approaches is what feature to use. In order to evaluate the quality of different features, we train Pre-act ResNet 32 for one episode on CIFAR-10. Then we obtain t-SNE 2D embeddings from the outputs of three different layers as shown in Fig.~\ref{fig:deep representation}. 

As can be seen, features extracted from shallow layers are mixed together regardless of their classes. Conversely, features extracted from the last convolutional layer and the fully connected layer are clustered together if they belong to a same class, and separated from each other if they are from different classes. This observation indicates deeper features are more useful than shallow features for KNN, which is consistent with the curve shown in Fig.~\ref{fig:deep_feature}. Meanwhile, the recovery rate of using the last convolution layer as embeddings outperforms the others. We believe the fully connected layer is not suitable for label recovery either, because it is directly related to the logit layer, which is highly likely to be corrupted during training on noisy dataset.

%%%%%%%%%%%%%%%%%%%%%%%%%
\begin{comment}
\begin{table}[h!]
\setlength{\abovecaptionskip}{-0.3cm}
\begin{center}
\begin{tabular}{c|cccc}
\hline
 methods &  \multicolumn{4}{c}{asymmetric noise} \\
  & 10 & 20 & 30 & 40  \\
\hline\hline 
 Joint Optimization & 96.35 & 95.12 & 93.96 & 88.16 \\
Pencil & 96.36 & 95.24 & 93.31 & 88.52 \\
IterKNN & \textbf{94.47} & 95.77 & 93.04 & 88.92 \\
SelKNN & 93.29 & \textbf{96.01} &  \textbf{94.15} & \textbf{90.03} \\
\hline
\end{tabular}
\end{center}
\caption{Label Recovery Comparison on CIFAR-10 with different asymmetric noise levels}\label{table:label_recovery_asym}
\end{table}
\end{comment}

%%%%%%%%%%%%%%%%%%%%%%%%%%%%%%%%%%%%%%%%%%%%%%%%%%%%%%%%%%%%%%%%%%%%%%%%%%%%%%%%%
\begin{table}[h!]
%\scriptsize
\setlength{\abovecaptionskip}{-0.2cm}
\begin{center}
\begin{tabular}{c|ccccc}
\hline
 methods & CE & Joint Opt & Pencil & SelKNN \\
 \hline
 accuracy (\%) & 68.8  & 72.23 & 73.49 & \textbf{76.78}   \\
\hline
\end{tabular}
\end{center}
\caption{Compare proposed method with baselines on Clothing1M.}\label{table:Clothing1M}
\end{table}

\subsection{Experiments on Real-world Noisy Dataset}
Finally, we test our approach on Clothing1M database~\cite{xiao2015learning}, a large real-world dataset composed of clothing images crawled from online shopping websites. Clothing1M comprises 1 million images with real noisy labels with additional 48 thousands verified clean data for training. Its overall noise proportion is approximately $38 \%$. We adopt Resnet-50 pretrained on ImageNet as backbone. The data preprocessing procedure includes resizing the image with a short edge of 256 and randomly cropping a 224$\times$224 patch from the resized image.  We use the Adam optimizer and the weight decay factor is 0.005. The initial learning rate is 0.002 and decreased  by  10  every  5  epochs. The batch size is 64.

Since the dataset is very large, it is not practical to search KNN across all the data samples. Therefore we only apply SelKNN for label correction, where the top 10,000 images are selected from each class as reference. There are 14 classes, thus in every round of label correction, our KNN database has 140,000 images in total. We set $k=500$ and use L2 distance metric with majority voting for classification. We perform in total 5 training episodes and label updates, where each episode contains 15 epochs. The comparison with other 4 approaches is shown in Table~\ref{table:Clothing1M}. Our method achieves state-of-the-art accuracy and is more than $3\%$ higher than the second best method.

%cifar10, when 80 noise, can correct 87 labels, when 60 %?, when 40 ? when 20 ?
%Cifar 100 when 80 noise, can correct ? labels, when 60 %?, when 40 ? when 20 ?

%A figure on Cifar10 with noise 60 percent, on Cifar100 with noise 60 percent.

%\subsection{Distance Metrices: L2 vs Cos similarity} may also be in the supplementary

%\subsection{majority voting vs exponential distance} may be in supplementary

%\subsection{Combining with SCE loss} may be in supplementary

% \begin{table}
% \begin{center}
% \begin{tabular}{|c|cccc|ccc|}
% \hline
%  methods & \multicolumn{4}{c|}{symmetric noise} & \multicolumn{3}{c|}{asymmetric noise} \\
%   & 20 & 40 & 60 & 80 & 20 & 30 & 40  \\
% \hline\hline 
%  first Conv & 99.01 & 0 & 0 & 0 & 0 & 0 & 0\\
%  Last Conv & 99.05 & 0 & 0 & 0 & 0 & 0 & 0\\
% FC & 99.05 & 0 & 0 & 0 & 0 & 0 & 0\\

% \hline
% \end{tabular}
% \end{center}
% \caption{different layer as representation}
% \end{table}

%\begin{table}
%\begin{center}
%\begin{tabular}{|c|cccc|ccc|}
%\hline
% methods & \multicolumn{4}{c|}{symmetric noise} & \multicolumn{3}{c|}{asymmetric noise} \\
%  & 20 & 40 & 60 & 80 & 20 & 30 & 40  \\
%\hline\hline 
% L2+ weighted dst & 99.01 & 0 & 0 & 0 & 0 & 0 & 0\\
%cos + majority  & 99.05 & 0 & 0 & 0 & 0 & 0 & 0\\
%cos +weighted dst & 99.05 & 0 & 0 & 0 & 0 & 0 & 0\\
%\hline
%\end{tabular}
%\end{center}
%\caption{different distance metrices}
%\end{table}

\section{Conclusion}
In this paper, we propose \textit{k}-nearest neighbor based iterative label correction framework for learning on the corrupted dataset. The approach is effective due to the robustness nature of KNN. We apply iterative retraining after every round of label correction to escape from over-fitting. To further mitigate the impact of wrong labels, we use loss ranking to select clean samples as reference for KNN classification. We have conducted abundant experiments on both synthetic and real-world datasets. Empirical results show that our SelKNN algorithm achieves state-of-the-art performance. An interesting related direction is to study how to leverage deep KNN to defend backdoor attack~\cite{chen2018detecting}.
%For future work, it is worth exploring how our approach is affected by different clean sample filtering methods. 

%Another interesting direction is to study how to leverage deep KNN to defend backdoor attack~\cite{chen2018detecting}, which is a related but more challenging problem than learning on noisy dataset.

{\small
\bibliographystyle{splncs04}
\bibliography{main}
}

\end{document}

\large{\textbf{Appendix}}

\appendix

\section{asymmetric noise label recover comparison}

We compare the label recovery rate of our approach with baselines on CIFAR-10 with different asymmetric noise level as shown in table. We also compare the label recovery rate of our approach with baselines on CIFAR-100 with both different symmetric and asymmetric noise level as shown in table.

\section{different metrices}

In our algorithm and experiments, we always use L2, will cosine similarity help? we use majority voting to create hard labels? how about soft labels?

We weight equally all reference neighbors. What if weight them based on distance to the query image. Will it be more robust?

\section{Deep knn to make prediction on test dataset}

What if we throw out the layers after the deep representation, and maintain all the training sample labels(or filtered clean labels) together with corresponding deep representations. Will that be better? 

\begin{table}
\begin{center}
\begin{tabular}{c|cccc}
\hline
 methods &  \multicolumn{4}{c}{asymmetric noise} \\
  & 10 & 20 & 30 & 40  \\
\hline\hline 
 Joint Optimization & 96.35 & 95.12 & 93.96 & 88.16 \\
Pencil & 96.36 & 95.24 & 93.31 & 88.52 \\
IterKNN & \textbf{94.47} & 95.77 & 93.04 & 88.92 \\
SelKNN & 93.29 & \textbf{96.01} &  \textbf{94.15} & \textbf{90.03} \\
\hline
\end{tabular}
\end{center}
\caption{Label Recovery Comparison on CIFAR-10 with different asymmetric noise levels}\label{table:label_recovery_asym}
\end{table}